\pdfoutput=1

\documentclass[11pt]{article}

\usepackage{ACL2023}

\usepackage{times}
\usepackage{latexsym}

\usepackage[T1]{fontenc}

\usepackage[utf8]{inputenc}

\usepackage{microtype}

\usepackage{inconsolata}

\usepackage{times}
\usepackage{latexsym}
\usepackage{microtype}
\usepackage{import}
\usepackage{hyperref}
\usepackage{caption}
\usepackage{subcaption}
\usepackage{xcolor}
\usepackage{natbib}
\usepackage{latexsym,amsmath,amssymb,amsthm}
\usepackage{cleveref}
\usepackage{paralist}
\usepackage{sistyle}
\usepackage{tikz}
\usepackage{bbding}
\usepackage[normalem]{ulem}
\usepackage{tabularx}
\usepackage{booktabs}
\usepackage{makecell}
\usepackage{array}
\usepackage{tcolorbox}
\usepackage{multirow}
\usepackage{colortbl}

\definecolor{ForestGreen}{RGB}{34,139,34}

\title{Xwin-LM: Strong and Scalable Alignment Practice for LLMs}

\author{Bolin Ni$^{1,3}$, Jingcheng Hu$^{2,3}$, Yixuan Wei$^{2,3}$, \\
\textbf{Houwen Peng}$^{3,\star}$\textbf{,}   \textbf{Zheng Zhang}$^{3}$\textbf{,} \textbf{Gaofeng Meng}$^{1}$\textbf{,}  \textbf{Han Hu}$^{3}$ \vspace{0.7em} \\
{$^1$Institute of Automation, CAS.} \quad
{$^2$Tsinghua University} \quad
{$^3$Microsoft Research Asia}
 \\
\small{\texttt{nibolin2019@ia.ac.cn}} \quad
\small{\texttt{hujc22@mails.tsinghua.edu.cn}} \quad
\small{\texttt{\{t-yixuanwei, zhez, houwen.peng\}@microsoft.com}}
\\
\small{\texttt{gfmeng@nlpr.ia.ac.cn}} \quad
\small{\texttt{ancientmooner@gmail.com}} 
}

\begin{document}
\maketitle
\renewcommand{\thefootnote}{}
\footnotetext{$^\star$contact person}

\begin{abstract}
In this work, we present \textbf{Xwin-LM}, a comprehensive suite of alignment methodologies for large language models (LLMs). This suite encompasses several key techniques, including supervised finetuning (SFT), reward modeling (RM), rejection sampling finetuning (RS), and direct preference optimization (DPO). The key components are as follows: (1) \textbf{Xwin-LM-SFT}, models initially finetuned with high-quality instruction data; (2) \textbf{Xwin-Pair}, a large-scale, multi-turn preference dataset meticulously annotated using GPT-4; (3) \textbf{Xwin-RM}, reward models trained on Xwin-Pair, developed at scales of 7B, 13B, and 70B parameters; (4) \textbf{Xwin-Set}, a multiwise preference dataset in which each prompt is linked to 64 unique responses generated by Xwin-LM-SFT and scored by Xwin-RM; (5) \textbf{Xwin-LM-RS}, models finetuned with the highest-scoring responses from Xwin-Set; (6) \textbf{Xwin-LM-DPO}, models further optimized on Xwin-Set using the DPO algorithm. Our evaluations on AlpacaEval and MT-bench demonstrate consistent and significant improvements across the pipeline, demonstrating the strength and scalability of Xwin-LM. The repository \href{https://github.com/Xwin-LM/Xwin-LM}{\texttt{https://github.com/Xwin-LM}} will be continually updated to foster community research.
\end{abstract}

\section{Introduction}
Recent advances in artificial intelligence, epitomized by large language models (LLMs) such as GPT-4~\cite{gpt4} and Claude~\cite{claude3}, have demonstrated remarkable capabilities across diverse real-world applications. Ensuring these models align with human expectations and values is crucial, especially as they are integrated into and utilized across numerous applications~\cite{instructGPT,bai2022training}.

To achieve this alignment, the technique of Reinforcement Learning from Human/AI Feedback (RLHF/RLAIF)~\cite{rlhf, rlaif} has been proposed. This approach involves initially gathering preferences from human or AI sources, followed by optimizing a policy model against a clearly built Reward Model (RM)~\cite{instructGPT} or an implicit preference learning target~\cite{dpo}. While effective, the inherent complexity and high costs pose significant barriers, limiting extensive exploration within the research community.

\begin{figure} 
\centering
    \includegraphics[width=0.50\textwidth]{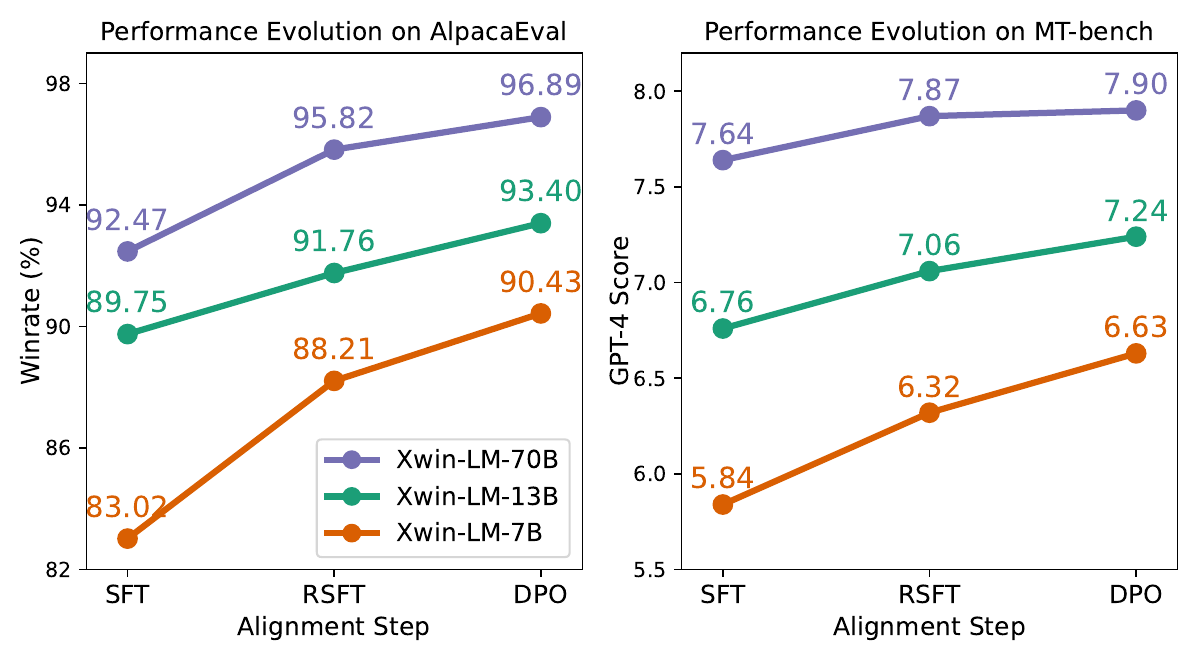}
    \caption{\textbf{Performance evolution.} The performance evolution on the AlpacaEval (left) and MT-bench (right) benchmarks suggests the strength and scalability for Xwin-LM, which can continuously improve the instruction-following ability on the 7B, 13B, and 70B scales. `SFT', `RSFT', and `DPO' denote supervised finetuning, rejection sampling finetuning, and direct preference optimization, respectively.} \label{fig:advertisement}
\end{figure}

In this work, we develop and release a strong and scalable RLHF pipeline named Xwin-LM. We detail our approach for developing Xwin-LM, which includes supervised finetuning, preference annotation, reward modeling, and policy optimization, along with observations and insights associated with each step. Specifically, we start with pretrained models Llama-2~\cite{llama2}, a collection of prompts, and a well-trained annotator, GPT-4. First, we train our supervised learning models, Xwin-LM-SFT, using an instruction-following dataset annotated by GPT-4 to establish an initial capability as a cold start. We then collect a preference dataset, Xwin-Pair, where responses are sampled from Xwin-LM-SFT and preferences are labeled by GPT-4. This dataset is used to train the reward model, Xwin-RM. Next, we use Xwin-LM-SFT to sample multiple responses for each prompt from another set and employ Xwin-RM to rank these responses, creating a multiwise preference dataset named Xwin-Set. Subsequently, Xwin-LM-RS is obtained by applying the rejection sampling (RS) finetuning technique to the highest-scoring responses in Xwin-Set. Finally, beyond learning only from positive samples, Xwin-LM-DPO employs the direct preference optimization technique, involving negative samples in Xwin-Set to learn from unexpected behavior.

We evaluate Xwin-LM on two popular instruction-following benchmarks, AlpacaEval~\cite{alpacaeval} and MT-bench~\cite{vicuna}. Fig.~\ref{fig:advertisement} illustrates the performance evolution of Xwin-LM throughout our pipeline. It is evident that Xwin-LM-SFT achieves a satisfactory cold start, and subsequent rejection sampling finetuning and direct preference optimization steps significantly improve model performance, indicating the strength of the proposed pipeline. Our Xwin-LM achieves state-of-the-art performance among all Llama2-based models.

In addition to the strong results, we have also gleaned several observations and insights associated with the pipeline: 
\begin{enumerate}[(1)] 
    \item \textbf{The model's upper capability limit remains fairly constant during RLHF; performance gains are mainly due to enhanced stability in generating high-quality responses.} Specifically, we observe that performance on two benchmarks and the RM score on our validation set under the best-of-1 protocol improved steadily, while those under the best-of-64 protocol remained fairly constant. 
    \item \textbf{For SFT, a linear enhancement in performance hinges on an exponential increase in data scale.} Furthermore, as the data scale continues to increase, performance gradually approaches saturation. 
    \item \textbf{Best-of-n evaluation is a discriminative metric for evaluating RMs and can also be an indicator for probing the potential optimization upper bound for alignment.} 
    \item \textbf{The DPO algorithm shows a certain sensitivity to the dispreferred responses within the data pair.} We find that the dispreferred response should closely match the policy's output distribution. 
\end{enumerate}

\section{Overview of XwinLM}
\begin{table}[t!]
    \small
    \centering
    \setlength{\tabcolsep}{2pt}
    \begin{tabular}{lrrrrr}
        \toprule
        Dataset & \makecell[c]{Num.\\of\\ conv.} & \makecell[c]{\#Turn\\ per\\ conv.} & \makecell[c]{\#Token\\ per\\ conv.} & \makecell[c]{\#Token\\ per\\ prompt} & \makecell[c]{\#Token\\ per\\ response} \\
        
        \midrule
        \multicolumn{6}{l}{\textbf{Step 1: SFT}} \\
        ShareGPT-Part-\uppercase\expandafter{\romannumeral1} & 6,206 & 7.3 & 4504.8 & 156.8 & 462.9 \\
        \midrule
        \multicolumn{6}{l}{\textbf{Step 2: RM}} \\
        ShareGPT-Part-\uppercase\expandafter{\romannumeral2} & 29,565 & 3.3 & 1634.7 & 67.6 & 414.1 \\
        Evo-Instruct-V2 & 142,992 & 1.0 & 733.1 & 145.0 & 572.8 \\
        \midrule
        \multicolumn{6}{l}{\textbf{Step 3: RSFT \& Step 4: DPO}} \\
        ShareGPT-Part-\uppercase\expandafter{\romannumeral3} & 39,861 & 2.4 & 1811.5 & 108.7 & 501.9 \\

        \bottomrule
    \end{tabular}
    \caption{\textbf{Statistics of datasets.} The prompts are from ShareGPT and Evo-Instruc-V2. We split the conversations in ShareGPT into three disjoint parts. The responses used in step 1 are from gpt-4, and the responses in step 2 and step 3 are from our Xwin-LM-SFT. The \#Turn and \#Token are averaged across all samples.}
    \label{tab:data_sources}
\end{table}
\subsection{High-level Methodology}
We begin with the pretrained LLM Llama-2~\citep{llama2}, a distribution of prompts, and well-trained AI annotators gpt-4. We then apply the following four steps.

\textbf{Step 1: Supervised Fine-Tuning (SFT).} We first finetune a pretrained Llama-2 on a demonstration dataset in a supervised fashion to obtain an initially aligned model. 

\textbf{Step 2: Collect comparison data, and train a reward model (RM).} We collect a dataset of comparisons
between model outputs, where annotators indicate which output they prefer. We then train a reward model to predict the quality of output.

\textbf{Step 3: Rejection Sampling (RS) finetuning.} For each prompt, we generate multiple responses from the fine-tuned model obtained in Step 1, and subsequently finetune models using the responses with the highest RM scores.

\textbf{Step 4: Direct Policy Optimization (DPO).} Building on the imitation of optimal responses in Step 3, DPO is utilized to further minimize the likelihood of suboptimal responses.

\subsection{Datasets}
\textbf{Source of prompts.} Our prompt dataset comprises ShareGPT~\cite{sharegpt} and Evo-Instruct-V2~\citep{wizardlm}. The responses from these datasets are only utilized during the SFT stage. Detailed statistics about the datasets are shown in Tab.~\ref{tab:data_sources}. It is important to note that our focus is on exploring an effective and scalable alignment pipeline rather than creating a model with the strongest performance. Therefore, we use a limited data source and do not employ other task-specific datasets.

\noindent \textbf{Annotator.} 
We use the GPT-4 API as the annotator since recruiting and training human annotators is time-consuming and expensive~\citep{instructGPT}. The annotators and evaluators used in this work are consistently GPT-4, ensuring the transferability of our pipeline.

\subsection{Evaluation}
To assess instruction-following capabilities, we utilize two widely recognized benchmarks:
\begin{itemize}
  \item AlpacaEval~\citep{alpacaeval} is a single-turn benchmark consisting of 805 questions across various topics, primarily focusing on helpfulness. Models are evaluated by GPT-4, and the definitive metric is the pairwise win rate compared to the text-davinci-003.
  \item MT-bench~\citep{vicuna} presents a two-turn evaluation with 160 questions covering eight diverse fields such as writing, reasoning, and mathematics. The model is required to not only provide an answer to the first question but also a subsequent, predefined follow-up. The responses are evaluated by GPT-4 on a scale from 1-10, and the model's overall score is averaged on all questions.
\end{itemize}
We use gpt-4-0613 API as the evaluator. These benchmarks have established human agreement metrics to ensure their reliability. For MT-bench, we observe fluctuations in the results; hence, we conduct the evaluation three times and report the median value.

\section{Step 1: Supervised Finetune}
We initiate our instruct-following pipeline with supervised finetuning (SFT) on a high-quality conversation dataset using pretrained models. The training loss is computed exclusively for tokens associated with the response segment.
\subsection{Experiment Setup}
\textbf{Dataset.}
We utilize a subset of ShareGPT containing 6,206 conversations responded to by GPT-4~\cite{openchat} for SFT. These long conversations are split into blocks with a maximum length of 4,096 tokens, resulting in a total of 10,605 conversations. We format the conversations following the Vicuna~\cite{sharegpt} guidelines.

\noindent \textbf{Implementation details.} 
We initialize our model using pretrained Llama-2. The models are finetuned for 3 epochs using the AdamW~\citep{adamw} optimizer. The AdamW optimizer’s hyperparameters are set as follows: $\beta_1=0.9$, $\beta_2=0.999$, $\epsilon=1\times10^{-8}$, and weight decay of $0$. We employ a cosine learning rate schedule with a maximum learning rate of $2\times10^{-5}$. The hyperparameters remain consistent among the 7B, 13B, and 70B models.

\subsection{Experiment Results and Analysis}
\begin{figure} 
\centering
    \includegraphics[width=0.45\textwidth]{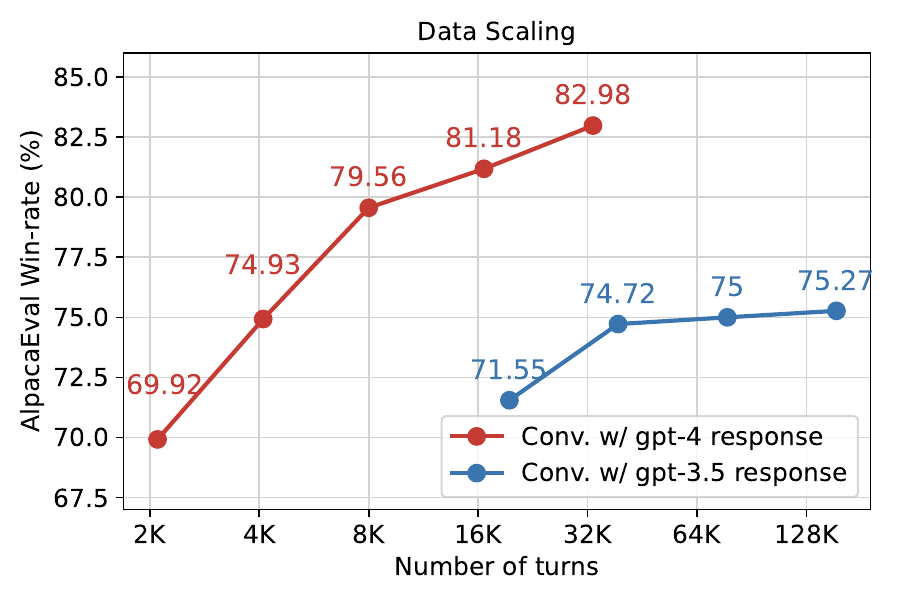}
    \caption{\textbf{Data scaling in SFT}. The performance is exponentially related to the data scale and gradually tends to saturate. The model trained on response from gpt-4 are significantly better than those from gpt-3.5-turbo.} \label{fig:data_scale}
\end{figure}
\textbf{Data quantity and quality.}  
We examine the effect of data quantity on performance using responses from both GPT-4 and GPT-3.5-turbo. Specifically, we sample conversations from ShareGPT-Part-I, with the number of turns ranging from 2k to 32k, and from ShareGPT-Part-II, with the number of turns ranging from 19k to 154k. The corresponding results are illustrated in Fig.~\ref{fig:data_scale}.

Initially, in the data incorporating GPT-4 responses, it is evident that a linear enhancement in performance necessitates an exponential increase in data quantity. Moreover, the acceleration in performance gains begins to decelerate when the number of turns surpasses 8k, indicating diminishing marginal utility in data quantity during the SFT stage.

Although the scope of this experiment is restricted due to the finite responses obtained from GPT-4, a similar trend is observed in the conversations incorporating GPT-3.5-turbo responses. An increase in data quantity from 38k to 154k results in only a marginal improvement of +0.45\%.

Furthermore, upon comparing models trained on different response sources, it is obvious that the performance of models trained on GPT-4 responses significantly outperforms those trained on GPT-3.5-turbo responses. For example, using 16k turns from GPT-4 outperforms 19k turns from GPT-3.5-turbo by a remarkable +9.63\% win rate. We conjecture that there are two potential explanations: 1) data quality holds more significance than quantity in SFT, and 2) the GPT-4 judge in AlpacaEval may favor models finetuned on its outputs.

\begin{figure*}[t!]
\centering
    \includegraphics[width=0.7\textwidth]{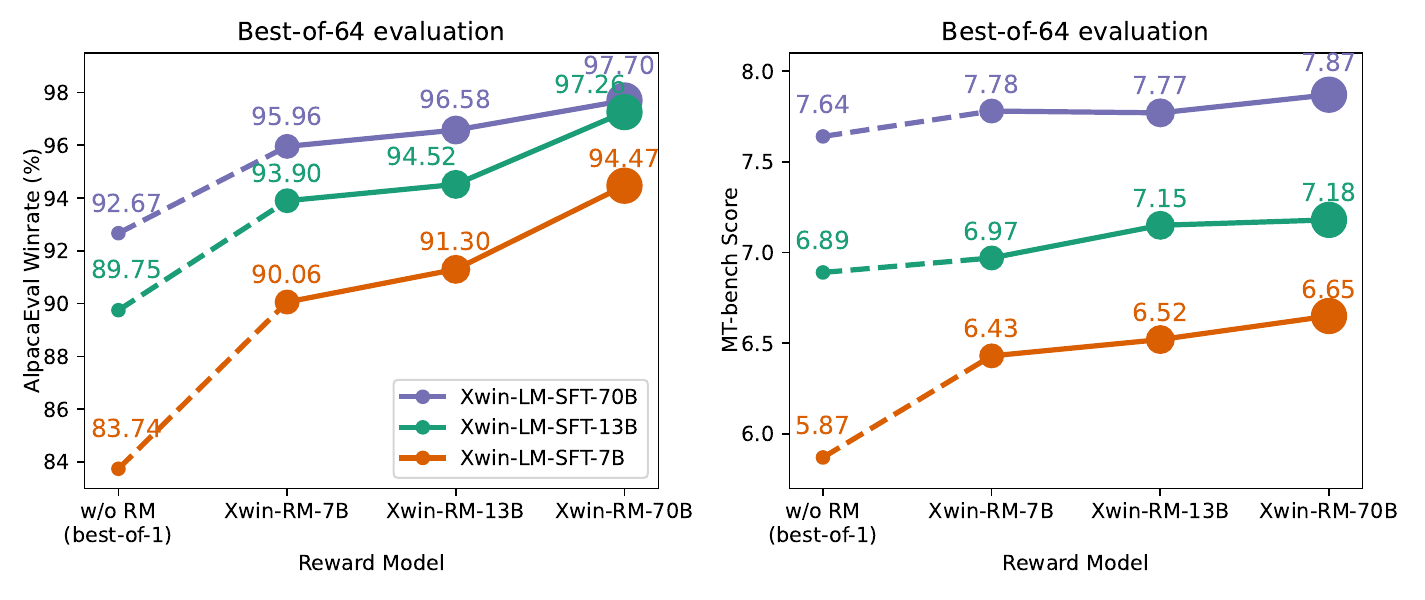}
    \caption{\textbf{Comparing RMs on best-of-64 evaluation protocol.} 
    Compared to sampling a single response, employing RM enables the selection of high-quality responses from a pool of 64 candidates, and larger RMs can select responses with superior performance, indicating that the capabilities of RM increase with size.} \label{fig:rm_best_of_n_v2}
\end{figure*}
\begin{figure*}[t!]
\centering
    \includegraphics[width=0.9\textwidth]{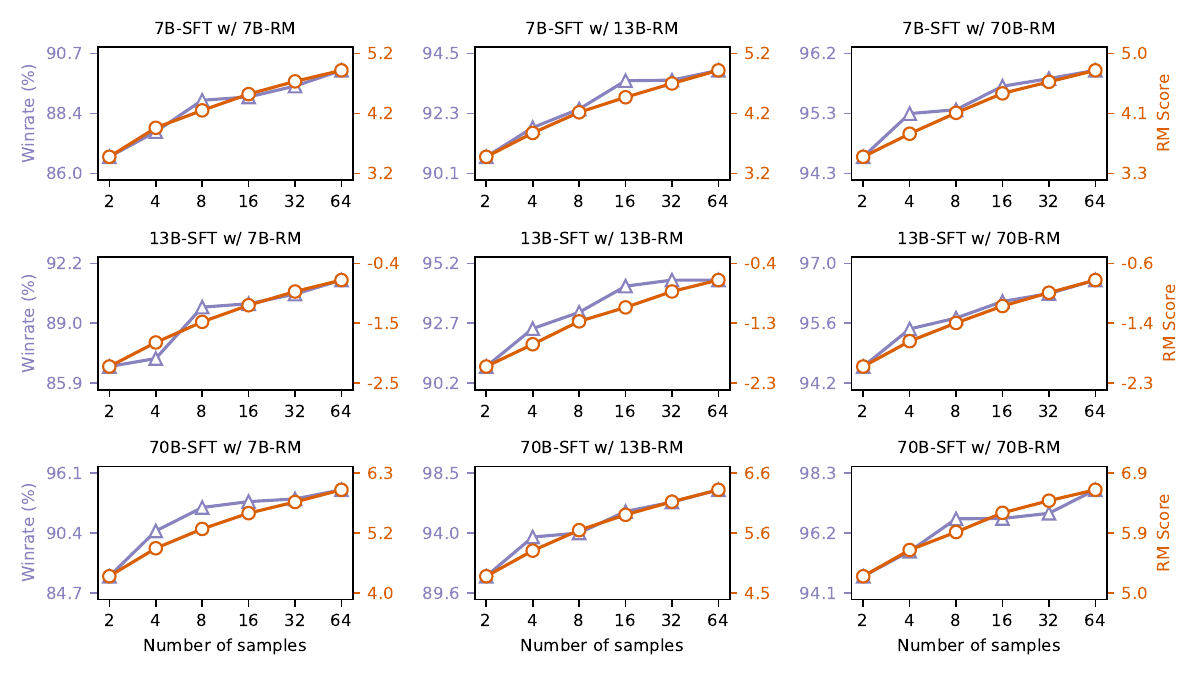}
    \caption{\textbf{Best-of-n evaluation.} We select the best response from different numbers of samples to evaluate the alignment between Xwin-RMs and the off-the-shelf AI judge. Left y-axis: AlpacaEval winrate judged by gpt-4. Right y-axis: RM score predicted by Xwin-RM-70B.} \label{fig:rm_best_of_various_n}
\end{figure*}

\begin{table}[t!]
  \centering
  \small
  \setlength{\tabcolsep}{2pt}
  \begin{tabular}{lccccc}
    \toprule
    &  \multirow{2}{*}{\shortstack{Significantly \\ Better}} & \multirow{2}{*}{Better} & \multirow{2}{*}{\shortstack{Slightly \\ Better}} & \multirow{2}{*}{\shortstack{Negligibly \\ Better/Unsure}} & \multirow{2}{*}{Total} \\
    & &&&& \\
    \midrule[\heavyrulewidth]
    Train & 35,596 & 57,832 & 124,638 & 9,843 & 227,909 \\
     & (16\%) & (25\%) & (55\%) & (4\%) & (100\%) \\
    \midrule
    Eval & 1,834 & 2,986 & 6,618 & 557 & 11,995 \\
     & (15\%) & (25\%) & (55\%) & (4\%) & (100\%) \\
    \bottomrule
  \end{tabular}
  \caption{\textbf{Statistic of Xwin-Pair.} The ``significantly better" category constitutes 15\% of the total data, while the majority of the data falls into the ``better" and ``slightly better" categories, accounting for 80\%.}
  \label{tab:rm_statistic}
\end{table}

\begin{table}[t!]
  \centering
  \small
  \setlength{\tabcolsep}{1pt}
  \begin{tabular}{lccccc}
    \toprule
    &  \multirow{2}{*}{\shortstack{Significantly \\ Better}} & \multirow{2}{*}{Better} & \multirow{2}{*}{\shortstack{Slightly \\ Better}} & \multirow{2}{*}{\shortstack{Negligibly \\ Better}} & \multirow{2}{*}{Avg} \\
    & &&&& \\
    \midrule[\heavyrulewidth]
    Xwin-RM-7B  & 77.71 & 70.56 & 65.91 & 82.51 & 69.45 \\
    Xwin-RM-13B & 78.25 & 70.06 & 66.03 & 62.80 & 69.52 \\
    Xwin-RM-70B & 71.48 & 82.93 & 72.23 & 67.34 & 71.48 \\
    \bottomrule
  \end{tabular}
  \caption{\textbf{Granular binary accuracy of Xwin-RM on our validation set.} Generally, reward models exhibit higher accuracy on more distinct responses (e.g., significantly better) and lower accuracy on similar responses (e.g., negligibly better).}
  \label{tab:rm_per_rating_acc}
\end{table}

\begin{table}[h]
\centering
\resizebox{.95\linewidth}{!}{
    \begin{tabular}{@{}lcc@{}}
    \toprule
    \multicolumn{1}{c}{Model} & AlpacaEval (\%) & MT-Bench \\ 
    \midrule 
    \midrule
    \multicolumn{3}{c}{\textit{Closed-source models}} \\
    GPT-4-1106 (Turbo)~\cite{gpt4} & 97.7 & 9.32 \\
    GPT-4-0613~\cite{gpt4} & 93.8 & 9.18 \\
    GPT-4-0314~\cite{gpt4} & 94.8 & 8.96 \\
    Claude-2~\cite{claude2} & 91.4 & 8.06 \\
    GPT-3.5-Turbo~\cite{gpt35turbo} & 93.4 & 8.39 \\
    \midrule
    \multicolumn{3}{c}{\textit{Open-source models LLaMA-2-7B}} \\
    Llama-2-7B-chat~\cite{llama2} & 71.4 & 6.27 \\
    Vicuna-7B-v1.5~\cite{vicuna} & - & 6.17 \\
    Tulu2 7B-DPO~\cite{tulu2} & 85.1 & 7.00 \\
    \cellcolor{gray!25}Xwin-LM-SFT-7B (ours) & \cellcolor{gray!25}\textbf{83.0} & \cellcolor{gray!25}\textbf{5.84} \\ 
    \cellcolor{gray!25}Xwin-LM-RS-7B (ours) & \cellcolor{gray!25}\textbf{88.2} & \cellcolor{gray!25}\textbf{6.32} \\ 
    \cellcolor{gray!25}Xwin-LM-DPO-7B (ours) & \cellcolor{gray!25}\textbf{90.4} & \cellcolor{gray!25}\textbf{6.63} \\ 
    \midrule

    \multicolumn{3}{c}{\textit{Open-source models LLaMA-2-13B}} \\
    Llama-2-13b-chat~\cite{llama2} & 81.1 & 6.65 \\
    Vicuna-13B-v1.5~\cite{vicuna} & - & 6.57 \\
    WizardLM-13B-v1.2~\cite{wizardlm} & 89.2 & 7.06 \\
    Tulu2 13B-DPO~\cite{tulu2} & 89.5 & 7.00 \\
    OpenChat 3.2 SUPER~\cite{openchat} & 89.5 & 7.19 \\
    \cellcolor{gray!25}Xwin-LM-SFT-13B (ours) & \cellcolor{gray!25}\textbf{89.8} & \cellcolor{gray!25}\textbf{6.76} \\ 
    \cellcolor{gray!25}Xwin-LM-RS-13B (ours) & \cellcolor{gray!25}\textbf{91.8} & \cellcolor{gray!25}\textbf{7.06} \\ 
    \cellcolor{gray!25}Xwin-LM-DPO-13B (ours) & \cellcolor{gray!25}\textbf{93.4} & \cellcolor{gray!25}\textbf{7.24} \\ 
    \midrule

    \multicolumn{3}{c}{\textit{Open-source models LLaMA-2-70B}} \\
    Llama-2-70b-chat~\cite{llama2} & 92.7 & 6.86 \\
    WizardLM-70B-v1.0~\cite{wizardlm} & 92.9 & 7.78 \\
    Tulu2 70B-DPO~\cite{tulu2} & 92.9 & 7.89 \\
    \cellcolor{gray!25}Xwin-LM-SFT-70B (ours) & \cellcolor{gray!25}\textbf{92.5} & \cellcolor{gray!25}\textbf{7.64} \\ 
    \cellcolor{gray!25}Xwin-LM-RS-70B (ours) & \cellcolor{gray!25}\textbf{95.8} & \cellcolor{gray!25}\textbf{7.87} \\ 
    \cellcolor{gray!25}Xwin-LM-DPO-70B (ours) & \cellcolor{gray!25}\textbf{96.9} & \cellcolor{gray!25}\textbf{7.90} \\ 
    \bottomrule
    \end{tabular}
}
\caption{\textbf{AlpacaEval and MT-bench results.} Xwin-LM achieves the SoTA performance step by step.}
\label{tab:performance}
\end{table}
\section{Step 2: Reward Modeling}
Our primary objective is to explore the RLHF pipeline rather than to seek the strongest RM. Therefore, we do not employ the existing preference datasets~\cite{ultrafeedback,bai2022training} in the community; 
instead, we build a preference dataset named Xwin-Pair, starting from a collection of real users' queries, and train Xwin-RM at 7B, 13B, and 70B scales based on it.

\noindent \textbf{Prompts collection.} 
First, we randomly sample 29,566 conversations from ShareGPT and also integrate the Evo-Instruct-V2 dataset~\cite{wizardlm} motivated by the scaling trends in Llama-2~\cite{llama2}, leading to an aggregate of 172,558 conversations. Given that the majority of conversations in the dataset span multiple turns, we unfold each conversation into multiple data instances by turn, with a length limit of 4,096 tokens. 

\noindent \textbf{Responses generation.} 
For each unfolded instance, we discard the response attached to the last turn in the original conversation. Instead, we use the conversation history and the query from the last turn as the prompt to the Xwin-LM-SFT to obtain the response. Sequentially, two distinct responses are sampled from Xwin-LM-SFT-70B at a temperature of 0.7 and 0.3, respectively. Both the top-$p$ and top-$k$ are set to 1.

\noindent \textbf{Preference annotation.} 
Then, we employ gpt-4-0314 API as a judge to provide three types of annotation for each instance: 1) which response is better; 2) the reason for this judgment; and 3) a rating among `significantly better', `better', `slightly better', and `negligibly better' following Llama-2. We randomly shuffle the two responses to avoid the judge's preference for the order of candidate responses. 

\noindent \textbf{Dataset statistics.}
We obtain a total of 239,904 preference data instances, splitting 227,909 instances for training and the remaining 11,995 for validation. More statistics are presented in Tab.~\ref{tab:rm_statistic}. We highlight that Xwin-Pair is the largest multi-turn preference dataset with additional explanations and fine-grained ratings. 

\noindent \textbf{Reward modeling.} Xwin-RM takes the response and its corresponding prompt (including the conversation history) as inputs and outputs a scalar score to indicate the quality of this response. We start from Xwin-LM-SFT and then add a randomly initialized linear head that outputs a scalar score. Specifically, given a prompt $x$, the training objective is to predict which response $y \in \{y_0, y_1\}$ is better. If the better response is $y_i$, we can write the loss function as:
\begin{equation}
    \mathcal{L}(r_\theta)=-\mathbb{E}_{\mathcal{D}_{RM}} [\log(\sigma(r_{\theta}(x,y_i)-r_{\theta}(x,y_{1-i})))]
\end{equation} 
where $r_\theta (x,y)$ is the scalar output of the reward model for prompt $x$ and response $y$ with parameters $\theta$, and $\mathcal{D}_{RM}$ is the preference dataset.

\subsection{Experiment Setup}
\textbf{Implementation Details.} We train for one epoch over the training set. The maximum learning rate is $2\times10^{-6}$ for the Xwin-RM-70B and $1\times10^{-5}$ for Xwin-RM-7B and 13B. The learning rate is decreased on a cosine learning rate schedule, down to 0. We use a warm-up of 3\% of the total number of steps. The effective batch size is 128 pairs for 70B RM and 256 for the rest.
 
\subsection{Experiment Results and Analysis}
\textbf{Granular binary accuracy.} 
The experimental results are presented in Tab.~\ref{tab:rm_per_rating_acc}, where there are two observations: 1) although more distinctive samples (e.g., `significantly better') account for only a small portion (15\%) of the training data, the reward models exhibit higher accuracy on these samples, yet lower accuracy on similar samples (e.g., `slightly better'), despite their higher data proportion (55\%). The influence of the proportion of training data with different levels of distinctiveness on the final performance of RM remains unclear, and we reserve this matter for future investigation; 2) larger RMs achieve higher average accuracy, but the improvement is not particularly dramatic. The accuracy only improved by 2.03\% when scaling the size of RM from 7B to 70B. We believe this is due to a combination of the inherent difficulty of this task and the instability of the preference annotation.

\noindent \textbf{Best-of-n evaluation as a practical indicator.} 
Beyond referencing binary accuracy, we further examine the generalizability of Xwin-RM by employing a best-of-n evaluation on AlpacaEval and MT-bench, which offer a diverse array of questions. For each question, the policy model generates 64 answers, after which Xwin-RM scores each answer, selecting the highest-scoring one for evaluation. Different RMs make their selections from the same pool of response candidates for the same policy. The results are depicted in Fig.~\ref{fig:rm_best_of_n_v2}. The three scales of Xwin-RMs are all capable of effectively selecting high-quality responses. Although the training data for RM includes only responses from the 70B SFT policy, Xwin-RMs also perform well on the 7B and 13B models, further demonstrating their generalizability. Additionally, while the smaller 7B RM can also select good answers for policies of varying scales, it is apparent that larger RMs can choose better responses than smaller ones, reaffirming the significance of RM model size.

\noindent \textbf{Xwin-RM aligns well with the off-the-shelf AI judge.}
To inspect the alignment between Xwin-RM and AI judge (e.g., gpt-4), we conduct the best-of-\{2,4,8,16,32,64\} evaluation on AlpacaEval, and compare the scores of RM and win-rate of the benchmark. The comprehensive results presented in Fig.~\ref{fig:rm_best_of_various_n} demonstrate a strict positive correlation between the scores of the Xwin-RM and the win-rates on the benchmark, indicating that Xwin-RMs can accurately score responses of varying quality, and their scoring outcomes are highly consistent with the judgments of GPT-4.

\begin{figure}[t]
\centering
    \includegraphics[width=0.48\textwidth]{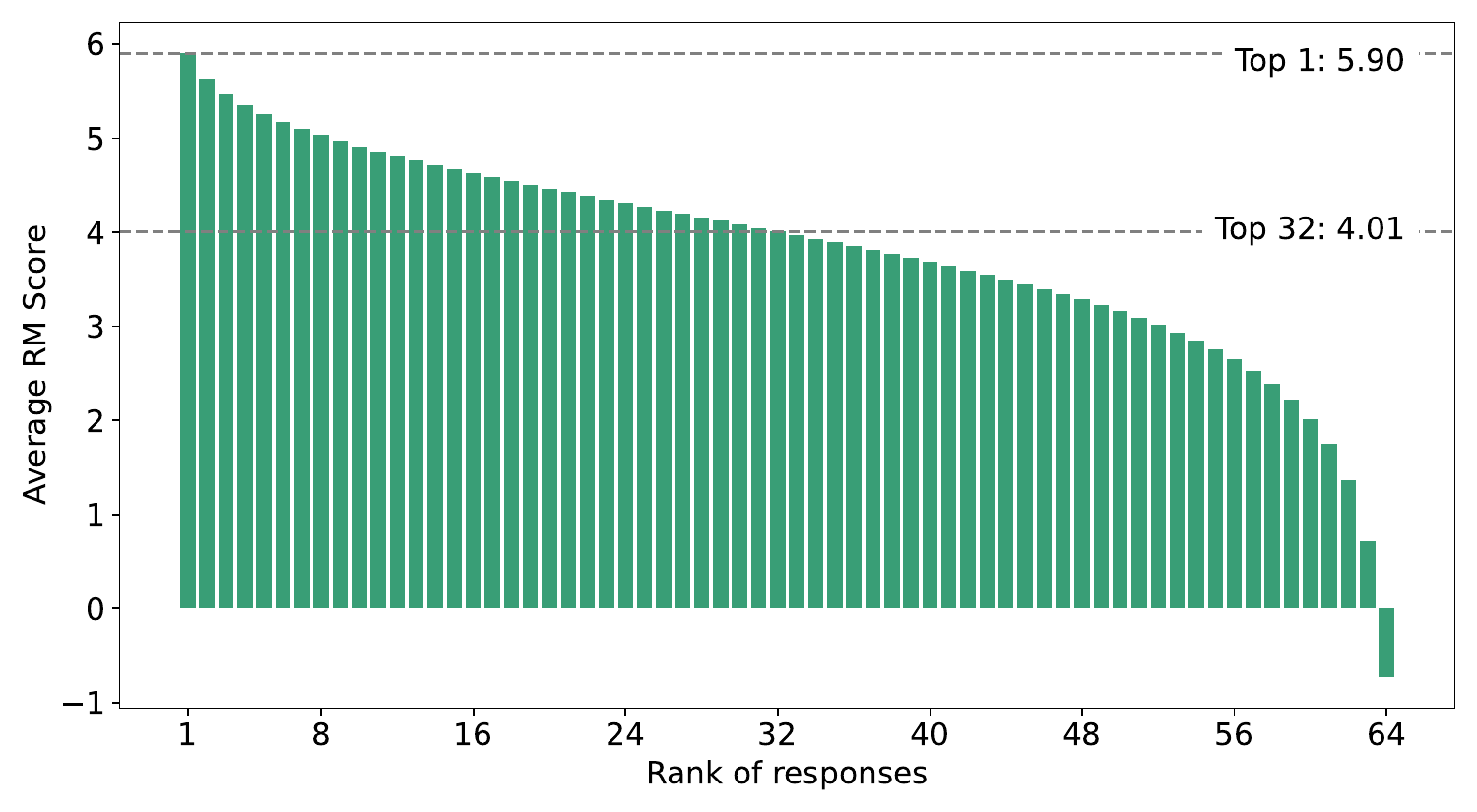}
    \caption{\textbf{Distribution of RM scores of responses at each rank.} Responses are sorted by score from highest to lowest, with the horizontal axis indicating the rank of the responses; a smaller rank signifies a higher score. The scores are averaged across all prompts.} \label{fig:score_per_rank}
\end{figure}
\begin{figure}[t] 
\centering
    \includegraphics[width=0.45\textwidth]{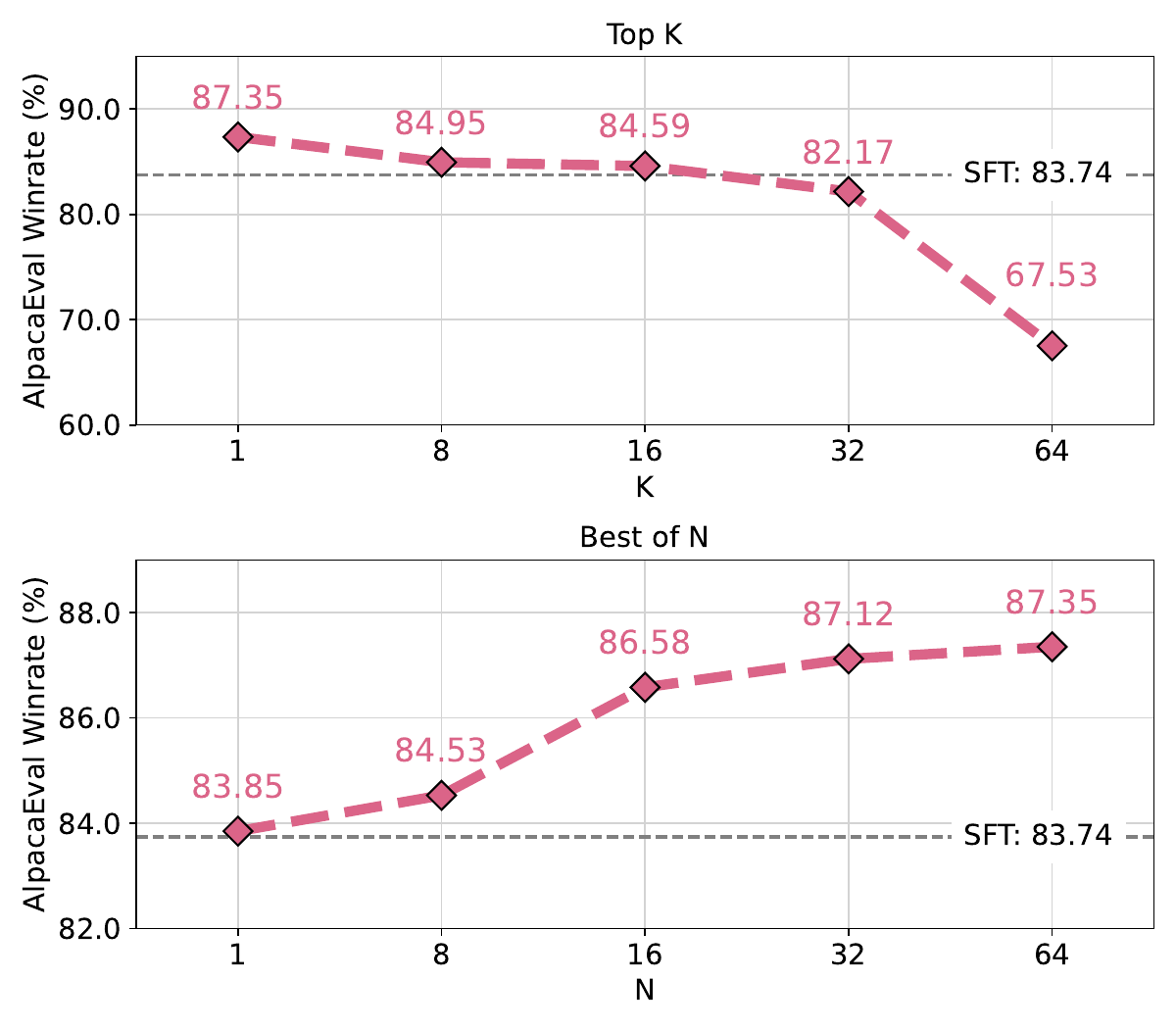}
    \caption{\textbf{Data selection in rejection sampling finetuning.} We explore the rank of selected samples (top) and the number of response candidates (bottom).} \label{fig:best_of_n_top_n}
\end{figure}

\begin{figure*}[t]
\centering
    \includegraphics[width=1\textwidth]{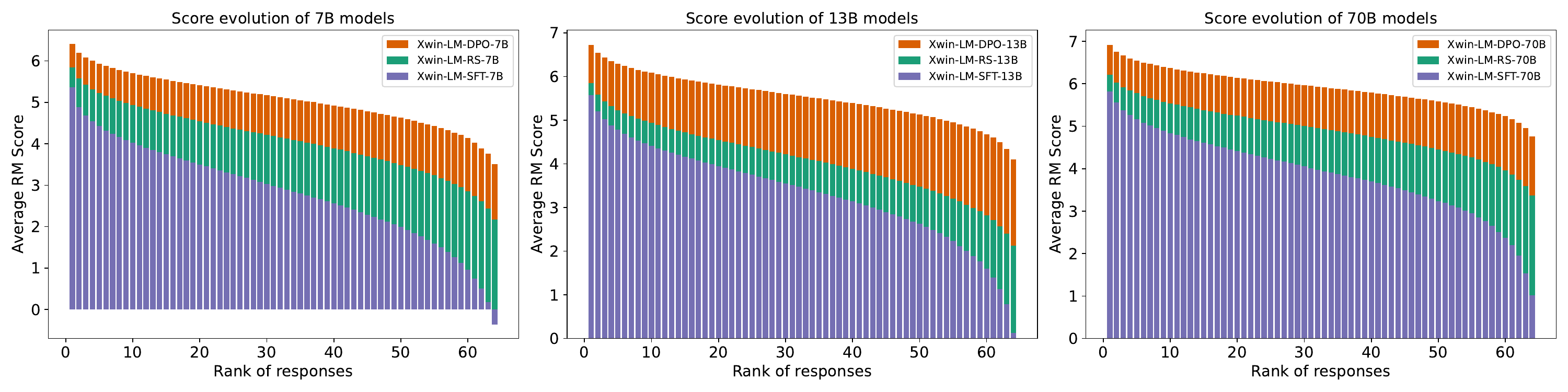}
    \caption{\textbf{Evolution of RM score with alignment pipeline.} With the alignment pipeline, there is a significant increase in the lower bound (e.g., responses at low rank) of the model's output scores, while the upper bound (e.g., responses at high rank) exhibits minimal change. This reflects that the enhancement in model performance is achieved by improving the stability of generating high-quality responses.} \label{fig:sft_rs_dpo}
\end{figure*}
\begin{figure}[t]
\centering
    \includegraphics[width=0.5\textwidth]{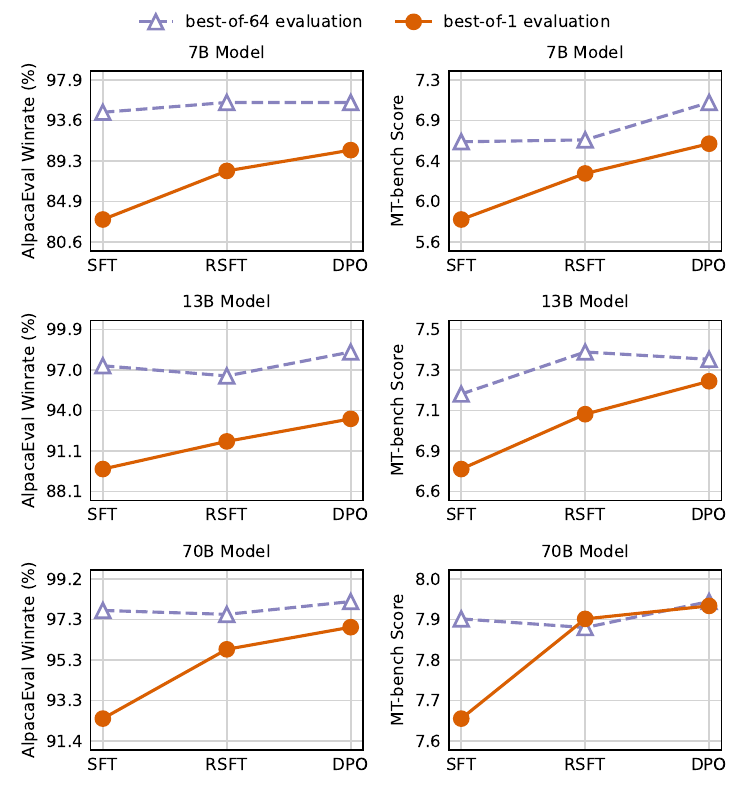}
    \caption{\textbf{Evolution of best-of-1 performance and best-of-64 performance.} 
    Although model performance consistently improves on the best-of-1 evaluation protocol, it exhibits a relatively steady trend on the best-of-64 evaluation. Left: AlpacaEval. Right: MT-bench.} \label{fig:upperbound}
\end{figure}

\section{Step 3: Rejection Sampling Finetune}
For each prompt, we sample N responses from the SFT policy model, and the response with the highest RM score is considered as the new ground truth. Subsequently, we finetune our model on these responses, thereby reinforcing the reward.

\subsection{Experiment Setup}
\textbf{Dataset construction.}
We randomly sample 27,424 ShareGPT conversations, distinct from the SFT and RM stages, and split them into 39,861 sub-conversations with a length limit of 4,096 tokens each. For each sub-conversation, we unfold the first three turns (if they exist) with the conversation history and generate 64 responses per last turn using Xwin-SFT-70B. These responses are then ranked with Xwin-RM-70B, yielding a multiwise preference dataset named Xwin-Set, which comprises a total of 96,277 prompts and 96,277 x 64 responses. Fig.~\ref{fig:score_per_rank} displays the average RM scores for all samples at each rank.

\noindent \textbf{Implementation details.} 
We employ the same hyperparameters as in the SFT stage, except for reducing the learning rate to $1\times10^{-5}$. Note that for each sub-conversation, we only compute the loss on the response of the last turn, as the responses in earlier turns are not generated by our model.

\subsection{Experiment Results and Analysis}
We conduct a preliminary investigation using Xwin-LM-RS-7B with half of the entire rejection sampling dataset.

\noindent \textbf{Influence of the rank of selected samples.}
The top panel of Fig.~\ref{fig:best_of_n_top_n} demonstrates that models trained with higher-ranked samples consistently achieve superior performance, corroborating the precision of our RM in ranking candidate responses.

\noindent \textbf{Effect of the number of response candidates.}
The bottom panel of Fig.~\ref{fig:best_of_n_top_n} reveals that increasing the number of response candidates enhances the quality of the samples obtained. However, the marginal gains begin to plateau beyond a sample size of 32. Doubling the number of samples from 32 to 64 leads to a modest win rate improvement of +0.23\%, which means that the sample size of 32 represents a great balance between computational efficiency and sample quality.

\section{Step 4: Direct Preference Optimization}
We employ DPO~\cite{dpo} instead of PPO~\cite{ppo} for two main reasons: 1) we observe that PPO is challenging to train on larger policy models, including hyperparameter issues and computational costs; 2) the rich preference relationships present in the Xwin-Set remain underutilized, whereas DPO inherently utilizes the preference data.

The DPO method directly updates the target policy from the preference pairs $(x, y_w, y_l)$ as follows:
\begin{footnotesize}
\begin{align}
    \mathcal L_\mathrm{DPO}&(\pi_\theta)=-\mathbb{E}_{(x,y_w,y_l)\sim\mathcal D}\\
    &
    \left[\log\sigma\left(\beta\left(
    \log\frac{\pi_\theta(y_w\mid x)}{\pi_\mathrm{ref}(y_w\mid x)}-
    \log\frac{\pi_\theta(y_l\mid x)}{\pi_\mathrm{ref}(y_l\mid x)}
    \right)\right)\right]\notag.
\end{align}
\end{footnotesize}The intuitive objective is to increase the likelihood of the preferred response $y_w$ and decrease the likelihood of the dispreferred response $y_l$. 

\begin{figure}[t]
\centering
    \includegraphics[width=0.5\textwidth]{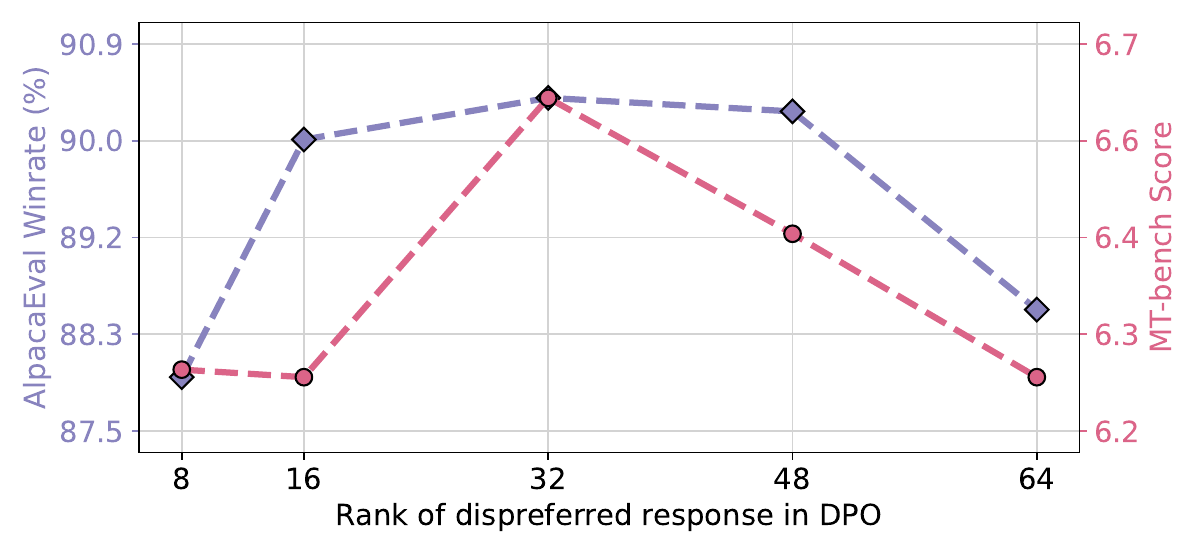}
    \caption{\textbf{Exploring the rank of dispreferred response.} The top-32 response is closest in score to that generated by Xwin-LM-RS-7B. Using either a higher or lower quality response as the dispreferred one impacts performance. Left y-axis: AlpacaEval. Right y-axis: MT-Bench.
} \label{fig:reject_selection}
\end{figure}

\noindent \textbf{Construction of preference pair.} 
For selecting preferred and dispreferred data, we follow the intuition that a model should learn from the highest quality data while correcting its most common mistakes. Based on this premise, we use the top-1 response as the preferred response. For the dispreferred response, a straightforward approach would be to have the policy model generate answers for all questions in the Xwin-Set. However, this method incurs additional generation costs. Therefore, we approximate the model's current capability using existing responses in the Xwin-Set. Specifically, we sample 500 prompts from the Xwin-Set, generate responses with the policy, and score them using Xwin-RM-70B. We then select the response from the Xwin-Set with a score closest to the policy model's score as the dispreferred response. The experiments in Fig.~\ref{fig:reject_selection} confirm our speculation.

\subsection{Experiment Setup}
\noindent \textbf{Dataset.} 
we utilize all prompts from the Xwin-Set. The preferred responses across all models correspond to the top-1 responses within the Xwin-Set. As for the dispreferred responses, we select top-32 for the 7B model, top-20 for the 13B model, and top-8 for the 70B model.

\noindent \textbf{Implementation details.} 
Xwin-LM-DPO is initialized from Xwin-LM-RS and trained for two epochs. We use a linear learning rate scheduler with a peak learning rate of 5e-7 and 100 warmup steps. The global batch size is 32 and $\beta$ is set to 0.1.

\subsection{Experiment Results and Analysis}

\textbf{The dispreferred data should closely match the policy's output distribution.} We use Xwin-LM-DPO-7B to explore the selection of dispreferred data. The top-32 response in Xwin-Set is the closest in score to that generated by Xwin-LM-RS-7B. We also try the responses with higher quality (e.g., top-{8, 16}) and lower quality (e.g., top-{48, 64}) as the dispreferred data. Fig.~\ref{fig:reject_selection} demonstrates that the model performs optimally on both AlpacaEval and MT-bench when the top-32 response serves as the dispreferred data. Both excessively high and low-quality responses as dispreference lead to performance degradation, correlating with the degree of quality deviation. This supports our hypothesis that the dispreferred response should mirror the policy model's output to correct the frequent errors. Additionally, lower-quality responses as dispreferred data outperform higher-quality ones, providing practical insights for DPO data curation.

\noindent \textbf{RLHF enhances the model's stability of generating quality answers, with the capability upper bound remaining relatively unchanged.} 
In Fig.~\ref{fig:upperbound}, we depict the performance evolution under the best-of-1 and best-of-64 evaluation during the alignment pipeline. In the best-of-64 evaluation, we generate 64 potential responses for each prompt and use Xwin-RM-70B to select the one with the highest RM score for evaluation. We observe a consistent enhancement in best-of-1 performance with deeper alignment, while best-of-64 performance remains largely unchanged. For instance, the 7B model's best-of-1 performance on AlpacaEval improves by +7.41\%, in contrast to a mere +1.05\% increase in best-of-64 performance. In addition to public benchmarks, we evaluate the model on a held-out set of 500 prompts from ShareGPT. Fig.~\ref{fig:sft_rs_dpo} displays the evolution in the RM score distribution of the generated responses. Notably, the top scores show no drastic changes, suggesting that the quality of the best responses does not significantly improve, whereas the scores of lower-quality responses experience a marked increase with alignment, significantly reducing the gap with top responses. This further underscores that the essence of RLHF lies in increasing the stability of quality output rather than elevating the capability upper bound, which remains relatively constant.

\noindent \textbf{Best-of-n performance can be viewed as the optimization upper bound of alignment.}
In Fig.~\ref{fig:upperbound}, we observe that after RLHF, the DPO model's best-of-1 performance closely approaches the SFT model's best-of-64 performance. This indicates that a simple SFT model is capable of producing quality answers, but cannot ensure their stability. Therefore, we can employ the best-of-n evaluation method to probe the upper limits of performance, serving as a guide for subsequent alignment efforts.

\section{Conclusion and Limitation}
In this work, we present Xwin-LM, a strong, and scalable alignment practice. Our pipeline includes supervised finetuning, reward modeling, rejection sampling finetuning, and direct preference optimization. Our model maintained the top-1 position on the AlpacaEval from September 2023 to November 2023. However, our work has several limitations: 1) we did not deliberately enhance the model's multi-turn capabilities, which may limit its practical user experience; 2) we utilized a limited data source, which could affect the model's overall performance; 3) we observed that the model suffers from hallucinations to some extent, which may be caused by training on self-generated data; 4) we noticed that the annotations and evaluations by GPT-4 exhibit a certain degree of instability.

\bibliography{custom}

\begin{thebibliography}{19}
\expandafter\ifx\csname natexlab\endcsname\relax\def\natexlab#1{#1}\fi

\bibitem[{Achiam et~al.(2023)Achiam, Adler, Agarwal, Ahmad, Akkaya, Aleman, Almeida, Altenschmidt, Altman, Anadkat et~al.}]{gpt4}
Josh Achiam, Steven Adler, Sandhini Agarwal, Lama Ahmad, Ilge Akkaya, Florencia~Leoni Aleman, Diogo Almeida, Janko Altenschmidt, Sam Altman, Shyamal Anadkat, et~al. 2023.
\newblock Gpt-4 technical report.
\newblock \emph{arXiv preprint arXiv:2303.08774}.

\bibitem[{Anthropic()}]{claude2}
Anthropic.
\newblock \href {https://www-cdn.anthropic.com/bd2a28d2535bfb0494cc8e2a3bf135d2e7523226/Model-Card-Claude-2.pdf} {Model card and evaluations for claude models}.

\bibitem[{Anthropic(2023)}]{claude3}
Anthropic. 2023.
\newblock \href {https://www-cdn.anthropic.com/de8ba9b01c9ab7cbabf5c33b80b7bbc618857627/Model_Card_Claude_3.pdf} {Model card and evaluations for claude models}.

\bibitem[{Bai et~al.(2022)Bai, Jones, Ndousse, Askell, Chen, DasSarma, Drain, Fort, Ganguli, Henighan et~al.}]{bai2022training}
Yuntao Bai, Andy Jones, Kamal Ndousse, Amanda Askell, Anna Chen, Nova DasSarma, Dawn Drain, Stanislav Fort, Deep Ganguli, Tom Henighan, et~al. 2022.
\newblock Training a helpful and harmless assistant with reinforcement learning from human feedback.
\newblock \emph{arXiv preprint arXiv:2204.05862}.

\bibitem[{Chiang et~al.(2023)Chiang, Li, Lin, Sheng, Wu, Zhang, Zheng, Zhuang, Zhuang, Gonzalez et~al.}]{sharegpt}
Wei-Lin Chiang, Zhuohan Li, Zi~Lin, Ying Sheng, Zhanghao Wu, Hao Zhang, Lianmin Zheng, Siyuan Zhuang, Yonghao Zhuang, Joseph~E Gonzalez, et~al. 2023.
\newblock Vicuna: An open-source chatbot impressing gpt-4 with 90\%* chatgpt quality.
\newblock \emph{See https://vicuna. lmsys. org (accessed 14 April 2023)}, 2(3):6.

\bibitem[{Cui et~al.(2023)Cui, Yuan, Ding, Yao, Zhu, Ni, Xie, Liu, and Sun}]{ultrafeedback}
Ganqu Cui, Lifan Yuan, Ning Ding, Guanming Yao, Wei Zhu, Yuan Ni, Guotong Xie, Zhiyuan Liu, and Maosong Sun. 2023.
\newblock Ultrafeedback: Boosting language models with high-quality feedback.
\newblock \emph{arXiv preprint arXiv:2310.01377}.

\bibitem[{Ivison et~al.(2023)Ivison, Wang, Pyatkin, Lambert, Peters, Dasigi, Jang, Wadden, Smith, Beltagy et~al.}]{tulu2}
Hamish Ivison, Yizhong Wang, Valentina Pyatkin, Nathan Lambert, Matthew Peters, Pradeep Dasigi, Joel Jang, David Wadden, Noah~A Smith, Iz~Beltagy, et~al. 2023.
\newblock Camels in a changing climate: Enhancing lm adaptation with tulu 2.
\newblock \emph{arXiv preprint arXiv:2311.10702}.

\bibitem[{Lee et~al.(2023)Lee, Phatale, Mansoor, Lu, Mesnard, Bishop, Carbune, and Rastogi}]{rlaif}
Harrison Lee, Samrat Phatale, Hassan Mansoor, Kellie Lu, Thomas Mesnard, Colton Bishop, Victor Carbune, and Abhinav Rastogi. 2023.
\newblock Rlaif: Scaling reinforcement learning from human feedback with ai feedback.
\newblock \emph{arXiv preprint arXiv:2309.00267}.

\bibitem[{Li et~al.(2023)Li, Zhang, Dubois, Taori, Gulrajani, Guestrin, Liang, and Hashimoto}]{alpacaeval}
Xuechen Li, Tianyi Zhang, Yann Dubois, Rohan Taori, Ishaan Gulrajani, Carlos Guestrin, Percy Liang, and Tatsunori~B. Hashimoto. 2023.
\newblock Alpacaeval: An automatic evaluator of instruction-following models.
\newblock \url{https://github.com/tatsu-lab/alpaca_eval}.

\bibitem[{Loshchilov and Hutter(2017)}]{adamw}
Ilya Loshchilov and Frank Hutter. 2017.
\newblock Decoupled weight decay regularization.
\newblock \emph{arXiv preprint arXiv:1711.05101}.

\bibitem[{OpenAI(2023)}]{gpt35turbo}
OpenAI. 2023.
\newblock \href {https://openai.com/blog/gpt-3-5-turbo-fine-tuning-and-api-updates} {Gpt-3.5 turbo fine-tuning and api updates}.

\bibitem[{Ouyang et~al.(2022)Ouyang, Wu, Jiang, Almeida, Wainwright, Mishkin, Zhang, Agarwal, Slama, Ray et~al.}]{instructGPT}
Long Ouyang, Jeffrey Wu, Xu~Jiang, Diogo Almeida, Carroll Wainwright, Pamela Mishkin, Chong Zhang, Sandhini Agarwal, Katarina Slama, Alex Ray, et~al. 2022.
\newblock Training language models to follow instructions with human feedback.
\newblock \emph{NeurIPS}, 35:27730--27744.

\bibitem[{Rafailov et~al.(2024)Rafailov, Sharma, Mitchell, Manning, Ermon, and Finn}]{dpo}
Rafael Rafailov, Archit Sharma, Eric Mitchell, Christopher~D Manning, Stefano Ermon, and Chelsea Finn. 2024.
\newblock Direct preference optimization: Your language model is secretly a reward model.
\newblock \emph{NeurIPS}, 36.

\bibitem[{Schulman et~al.(2017)Schulman, Wolski, Dhariwal, Radford, and Klimov}]{ppo}
John Schulman, Filip Wolski, Prafulla Dhariwal, Alec Radford, and Oleg Klimov. 2017.
\newblock Proximal policy optimization algorithms.
\newblock \emph{arXiv preprint arXiv:1707.06347}.

\bibitem[{Stiennon et~al.(2020)Stiennon, Ouyang, Wu, Ziegler, Lowe, Voss, Radford, Amodei, and Christiano}]{rlhf}
Nisan Stiennon, Long Ouyang, Jeffrey Wu, Daniel Ziegler, Ryan Lowe, Chelsea Voss, Alec Radford, Dario Amodei, and Paul~F Christiano. 2020.
\newblock Learning to summarize with human feedback.
\newblock \emph{NeurIPS}, 33:3008--3021.

\bibitem[{Touvron et~al.(2023)Touvron, Martin, Stone, Albert, Almahairi, Babaei, Bashlykov, Batra, Bhargava, Bhosale et~al.}]{llama2}
Hugo Touvron, Louis Martin, Kevin Stone, Peter Albert, Amjad Almahairi, Yasmine Babaei, Nikolay Bashlykov, Soumya Batra, Prajjwal Bhargava, Shruti Bhosale, et~al. 2023.
\newblock Llama 2: Open foundation and fine-tuned chat models.
\newblock \emph{arXiv preprint arXiv:2307.09288}.

\bibitem[{Wang et~al.(2023)Wang, Cheng, Zhan, Li, Song, and Liu}]{openchat}
Guan Wang, Sijie Cheng, Xianyuan Zhan, Xiangang Li, Sen Song, and Yang Liu. 2023.
\newblock Openchat: Advancing open-source language models with mixed-quality data.
\newblock \emph{arXiv preprint arXiv:2309.11235}.

\bibitem[{Xu et~al.(2023)Xu, Sun, Zheng, Geng, Zhao, Feng, Tao, and Jiang}]{wizardlm}
Can Xu, Qingfeng Sun, Kai Zheng, Xiubo Geng, Pu~Zhao, Jiazhan Feng, Chongyang Tao, and Daxin Jiang. 2023.
\newblock Wizardlm: Empowering large language models to follow complex instructions.
\newblock \emph{arXiv preprint arXiv:2304.12244}.

\bibitem[{Zheng et~al.(2023)Zheng, Chiang, Sheng, Zhuang, Wu, Zhuang, Lin, Li, Li, Xing et~al.}]{vicuna}
Lianmin Zheng, Wei-Lin Chiang, Ying Sheng, Siyuan Zhuang, Zhanghao Wu, Yonghao Zhuang, Zi~Lin, Zhuohan Li, Dacheng Li, Eric Xing, et~al. 2023.
\newblock Judging llm-as-a-judge with mt-bench and chatbot arena.
\newblock \emph{arXiv preprint arXiv:2306.05685}.

\end{thebibliography}
\bibliographystyle{acl_natbib}


\end{document}